\documentclass{article}
\usepackage{fontspec}
\usepackage{amsmath}

\usepackage{microtype}
\usepackage{graphicx}
\usepackage{subcaption}
\usepackage{booktabs} 

\usepackage{hyperref}
\usepackage[T1]{fontenc}
\usepackage{booktabs}
\usepackage{multirow}
\usepackage{graphicx} 

\usepackage[preprint]{icml2026}

\usepackage{amsmath}
\usepackage{amssymb}
\usepackage{mathtools}
\usepackage{amsthm}

\usepackage[capitalize,noabbrev]{cleveref}
\definecolor{mygreen}{RGB}{200,255,200}
\theoremstyle{plain}

\theoremstyle{definition}

\theoremstyle{remark}

\usepackage[textsize=tiny]{todonotes}

\icmltitlerunning{CTRL-RAG}

\begin{document}

\twocolumn[
  \icmltitle{CTRL-RAG: Contrastive Likelihood Reward Based Reinforcement Learning for Context-Faithful RAG Models}



  \icmlsetsymbol{equal}{*}

  \begin{icmlauthorlist}
    \icmlauthor{Zhehao Tan}{equal,antgroup}
    \icmlauthor{Yihan Jiao}{equal,antgroup}
    \icmlauthor{Dan Yang}{antgroup}
    \icmlauthor{Junjie Wang}{antgroup}
    \icmlauthor{Duolin Sun}{antgroup}
    \icmlauthor{Jie Feng}{antgroup}
    \icmlauthor{Xidong Wang}{antgroup}
    \icmlauthor{Lei Liu}{antgroup}
    \icmlauthor{Yue Shen}{antgroup}
    \icmlauthor{Jian Wang}{antgroup}
    \icmlauthor{Jinjie Gu}{antgroup}

  \end{icmlauthorlist}
  \icmlaffiliation{antgroup}{ANT GROUP, Med-AQ}

  \icmlcorrespondingauthor{Yihan Jiao}{jiaoyihan.yh@antgroup.com}
  \icmlcorrespondingauthor{Dan Yang}{luoyin.yd@antgroup.com}

  \icmlkeywords{Machine Learning, ICML}

  \vskip 0.3in
]



\printAffiliationsAndNotice{\icmlEqualContribution}

\begin{abstract}
With the growing use of Retrieval-Augmented Generation (RAG), training large language models (LLMs) for context-sensitive reasoning and faithfulness is increasingly important. Existing RAG-oriented reinforcement learning (RL) methods rely on external rewards that often \textbf{fail to evaluate document faithfulness}, and may misjudge similar answers in open-domain settings. In addition, there is no RAG-based self-reward mechanism. Moreover, although such a mechanism could in principle estimate answer confidence given documents, the absence of objective feedback in a self-judgment can cause “hallucination accumulation” and eventual model collapse. To tackle these issues, we propose a novel \textbf{"internal--external" hybrid} reward framework centered on a \textbf{Contrastive Likelihood Reward (CLR)}. CLR directly optimizes the log-likelihood gap between responses conditioned on prompts with and without supporting evidence. This encourages the model to extract relevant evidence and increases its confidence when grounded in a specific context. Experiments show that our method (used alone or combined with external correctness rewards) achieves strong performance on single-hop, multi-hop, vertical-domain, and faithfulness benchmarks. Our training code and models are coming soon. 
\end{abstract}

\section{Introduction}

With the burgeoning prominence of Retrieval-Augmented Generation (RAG), the enhancement of contextual capabilities has become a focal point in the post-training phase of Large Language Models (LLMs). 
\begin{figure*}[ht]
  \vskip 0.2in
  \begin{center}
    \centerline{\includegraphics[width=\textwidth]{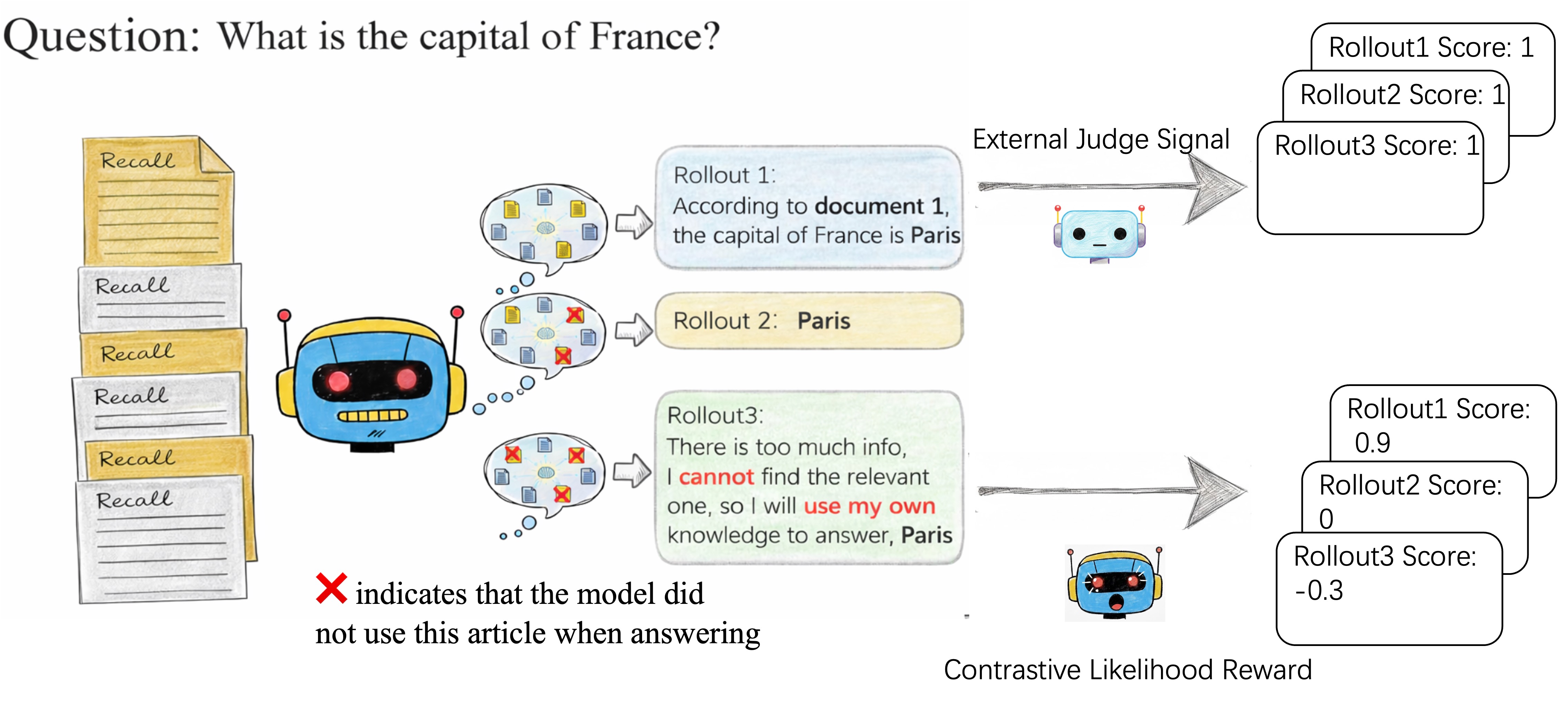}}
    \caption{
      The comparison between the traditional RAG RL methods (external judge signals) and our Contrastive Likelihood Rewards (CLR). All rollouts are generated using the same input; however, outcomes may vary depending on the extent to which the model utilizes the retrieved documents. A higher positive score indicates a greater degree of document utilization by the model, whereas a larger negative score indicates that the documents pose a greater burden on the model.
    }
    \label{fig:difference}
  \end{center}
\end{figure*}

Current approaches largely employ Supervised Fine-Tuning (SFT) to enhance these capabilities~\cite{Zhang_2024_03,jiao2025hiraghierarchicalthoughtinstructiontuningretrievalaugmented} or utilize Reinforcement Learning (RL) guided by external signals (e.g., accuracy, citation quality, format compliance)~\cite{huang2025ragrladvancingretrievalaugmentedgeneration, wu-etal-2025-pa}. However, these methods face some limitations:

\textbf{Deficiencies in External Evaluation to RAG Scenarios.} The external reward system is the mainstream approach for reinforcement learning in RAG-related scenarios, but it often acts as an imperfect judge of signals. For example, it fails to judge contextual faithfulness. Moreover, accuracy-oriented methods may yield false positives/negatives or fail to rank similar candidate answers effectively. Additionally, citation-based rewards are highly susceptible to noise from formatting errors or poor instruction adherence, thereby confounding the assessment of core reasoning proficiency as shown in Figure \ref{fig:difference}.

\textbf{Insufficient Adaptation and Attention of existing internal-rewarding methods to RAG Scenarios.} Existing uncertainty quantification and control methods based on entropy~\cite{zhang2025rightquestionhalfanswer} or perplexity~\cite{liu2025noverincentivetraininglanguage} are predominantly designed and evaluated in the context of general language modeling or open-ended generation tasks, and there is a lack of dedicated research targeting RAG, where answering strongly depends on external context. In RAG settings, models perform conditional generation given retrieved evidence, and both the sources of uncertainty and the way contextual information is exploited differ substantially from generation without such external inputs. However, current work has largely overlooked this scenario, lacking fine-grained characterization and systematic analysis of model answering behavior under contextual constraints. Moreover, Sole reliance on internal-rewards without objective external feedback, can lead to ``hallucination accumulation''~\cite{yuan2025selfrewardinglanguagemodels} in complex reasoning tasks, potentially causing model collapse. 

To address these challenges, we propose a novel reward mechanism that integrates intrinsic and extrinsic signals via contrastive likelihood estimation. In this framework, the intrinsic signal is derived from the log-probabilities of the generated answers, while the extrinsic supervision signal is sourced from pre-annotated supporting documents. This reward is specifically designed to bolster the model’s accuracy and reasoning capabilities given a query and retrieved context. Formally, we define the reward as the differential contrastive likelihood, calculated as the gap between the answer probability conditioned on both supporting and noisy documents versus the probability conditioned solely on noisy documents.

This signal can be used independently or superimposed with external correctness rewards. Our objective is twofold: (1) to enable the model to effectively identify the supporting document amidst noise. (2) To ensure that once the supporting document is located, the model exhibits increased confidence in generating the answer based strictly on that context.

Our contributions are summarized as follows:
\begin{itemize}
    \item \textbf{Novel RAG-Specific RL Framework.} We propose \textbf{CTRL-RAG}, the first reinforcement learning approach specifically designed to optimize contextual faithfulness and contextual reasoning in RAG scenarios using Contrastive Likelihood Rewards (CLR). By integrating intrinsic log-probabilities with extrinsic document supervision, CLR mitigates reward sparsity and ensures answers are grounded in retrieved evidence rather than parametric memory.
    \item \textbf{Robustness Across Architectures.} Extensive experiments on Dense and MoE models demonstrate significant performance gains, proving the effectiveness and generalizability of our hybrid reward mechanism.
\end{itemize}

\section{Related Work}
\textbf{Post-training for RAG Generator}. While RAG and DeepSearch systems enhance LLM knowledge and mitigate hallucinations\cite{zhang2023sirenssongaiocean, guu2020realmretrievalaugmentedlanguagemodel,lewis2021retrievalaugmentedgenerationknowledgeintensivenlp}, they demand generators with high faithfulness and context-aware reasoning for complex generation tasks. SFT-based methods like Self-RAG\cite{asai2023selfraglearningretrievegenerate} and HIRAG \cite{jiao2025hiraghierarchicalthoughtinstructiontuningretrievalaugmented}introduce specialized tokens or hierarchical data to improve evidence evaluation and information extraction. Meanwhile, RL approaches such as RAG-RL\cite{huang2025ragrladvancingretrievalaugmentedgeneration} and PA-RAG\cite{wu-etal-2025-pa} optimize models for citation accuracy and document preference. To evaluate these capabilities, benchmarks such as MuSiQue\cite{trivedi2022musiquemultihopquestionssinglehop} focus on reasoning, while PRGB\cite{tan2025prgbbenchmarkrobustplaceholderassisted} and FaithfulEval\cite{ming2025faithevallanguagemodelstay} specifically measure how well models prioritize external documents over internal priors through counterfactual or isolated testing.

\textbf{External and Internal Reward Signals}. Reinforcement Learning with Verifiable Rewards (RLVR) has excelled in deterministic domains like math and code\cite{Guo_2025,openai2024openaio1card,hu2025openreasonerzeroopensourceapproach,lambert2025tulu3pushingfrontiers}, yet applying it to general RAG tasks remains challenging. Rule-based rewards in RAG often lead to reward hacking, where models mimic formats without improving reasoning, while LLM-as-a-judge or Reward Models incur high costs and provide sparse, noisy signals\cite{Guo_2025,zhang2025surveyreinforcementlearninglarge,liu2025judgejudgeimprovingevaluation}. Conversely, internal signals—such as entropy\cite{zhang2025rightquestionhalfanswer, liu2025noverincentivetraininglanguage} and self-certainty\cite{zhao2025learningreasonexternalrewards} or self-rewarding mechanisms—leverage the model’s own confidence to improve accuracy\cite{yuan2025selfrewardinglanguagemodels}. However, internal signals alone cannot expand a model’s knowledge boundaries or ensure grounding in retrieved context. This gap suggests that a hybrid approach, combining external verification with internal confidence signals, is necessary to achieve robust faithfulness and reasoning.

\section{Preliminaries}
\label{sec:preliminaries}

In this section, we establish the theoretical framework for our reward design. We first formalize the generation process and the current rule-based reward mechanisms in Retrieval-Augmented Generation (RAG), and subsequently introduce the Group Relative Policy Optimization (GRPO) algorithm.

\subsection{Rewards for RAG Generator}
\label{subsec:pre_rag_rewards}

The RAG generator aims to produce a reliable and accurate response $y$ conditioned on a user query $q$ and a set of retrieved documents $\mathcal{D} = \{d_1, d_2, \dots, d_k\}$. Formally, the generator, parameterized by $\theta$, defines a policy $\pi_\theta$ that models the conditional probability:
\begin{equation}
P(y|q, \mathcal{D}) = \prod_{t=1}^T P(y_t | y_{<t}, q, \mathcal{D})
\end{equation}
where $T$ denotes the sequence length. In many open-domain QA scenarios, the quality of a generated response is typically evaluated using the following rule-based metrics:

\begin{itemize}
    \item \textbf{Citation Reward ($R_{cite}$):} This reward evaluates the attribution accuracy to encourage the model to ground its response in the retrieved context. Specifically, it verifies the presence of predefined citation markers, denoted as $c_i$ (e.g., ``[doc $i$]''), which correspond to the set of supporting documents $D^+ \subseteq D$. Formally, $R_{cite}$ is defined as the recall of required citations:
    \begin{equation}
        R_{cite} = \frac{1}{|\mathcal{D}^+|} \sum_{d_i \subseteq \mathcal{D}^+} \mathbb{I}(c_i \in y)
    \end{equation}
    where $\mathbb{I}(\cdot)$ is the indicator function that outputs 1 if the marker $c_i$ is present in the generated response $y$, and 0 otherwise.

    \item \textbf{Correctness Reward ($R_{acc}$):} When a ground-truth answer $y^*$ is available, $R_{acc}$ quantifies the semantic or exact matching between $y$ and $y^*$. In open-domain tasks, a common heuristic is to check if the ground-truth string $y^*$ is an exact substring of the response $y$:
    \begin{equation}
    R_{acc} = \mathbb{I}(y^* \subseteq y)
    \end{equation}

\end{itemize}

The total reward is generally formulated as a weighted linear combination:
\begin{equation}
R_{\text{total}} = \alpha R_{cite} + \beta R_{acc}  - \eta \cdot \text{Cost}(y)
\label{total}
\end{equation}
where $\alpha$, $\beta$ are scaling coefficients and $\text{Cost}(y)$ serves as a regularizer for sequence length or repetition.

\subsection{Group Relative Policy Optimization (GRPO)}
\label{subsec:grpo}

To optimize the RAG generator while addressing the stability issues of RL, we utilize the Group Relative Policy Optimization (GRPO) algorithm. For each
question q, the model rolls out a group of completions $\{o_1, o_2, \dots, o_G\}$. The objective is formulated as:
\begin{align}
    \mathbb{J}(\theta) &= \mathbb{E}_{q, \{o_i\}} \left[ \frac{1}{G} \sum_{i=1}^G \left( \frac{1}{|o_i|} \sum_{t=1}^{|o_i|} r_{i,t}^{clip}(\theta) - \beta \mathbb{D}_{\text{KL}} \right) \right] \label{eq:total_loss} \\[0.5ex]
    r_{i,t}^{clip}(\theta) &= \min \left( r_{i,t}(\theta) \hat{A}_i, clip(r_{i,t}(\theta), 1-\epsilon, 1+\epsilon) \hat{A}_i \right) \label{eq:clip_loss}
\end{align}
where $\mathbb{D}_{KL} = KL(\pi_\theta \| \pi_{ref})$ and $r_{i,t}(\theta)$ represents the token-level probability ratio:
\begin{equation}
r_{i,t}(\theta) = \frac{\pi_\theta(o_{i,t} | q, o_{i,<t})}{\pi_{\theta_{old}}(o_{i,t} | q, o_{i,<t})}
\end{equation}

The advantage $\hat{A}_i$ is computed by normalizing the total reward $R_{\text{total}, i}$ within the group:
\begin{equation}
    \hat{A}_{i,t} = \frac{R_{i, t} - \mu_R}{\sigma_R + \epsilon}
\end{equation}
where $\mu_R$ and $\sigma_R$ are the mean and standard deviation of rewards in the group.

\section{Methodology}

In the design of rewards for RAG generators, existing approaches predominantly rely on external reward models. However, these external signals are prone to reward hacking and frequent misjudgments, such as false positives. Furthermore, they often overlook the internal state and confidence signals of the generator itself. To address these limitations, we propose a hybrid reward mechanism that integrates both internal and external signals. By leveraging contrastive likelihood estimation, we define the reward as the marginal evidential contribution attributed to supporting documents. This approach effectively enhances the model's ability to filter out noise, encourages the strategic utilization of relevant documents, and ultimately improves the faithfulness of the generated responses.

\subsection{Evidential Contribution}
To quantify how much a generated response is grounded in the provided documents, we introduce a metric termed \textbf{Evidential Contribution}. This metric assesses the contribution of supporting documents to the model's confidence in generating a response. We adopt this specific terminology to distinguish our metric from the classical concept of Information Gain used in information theory and decision tree learning~\cite{quinlan1986induction}.

Let $q$ denote the question and $\mathcal{D} = \{d_1, d_2, \dots, d_n\}$ be a set of $n$ retrieved documents. We partition $\mathcal{D}$ into a subset of supporting documents, $\mathcal{D}^+$, which contain information relevant to answering $q$, and a subset of noisy or irrelevant documents, $\mathcal{D}^-$. Given a generated sequence (rollout) $y = (y_1, \dots, y_T)$ of length $T$ in which $y_1$ denotes the first token of $y$, we first define the sequence-level log-likelihood under the full context, $S(y \mid \mathcal{D})$, and its leave-one-out (LOO) counterpart, $S^{-}(y \mid \mathcal{D})$:
\begin{align}
    S(y \mid \mathcal{D}) &= \sum_{t=1}^{T} \log P(y_t \mid y_{<t}, q, \mathcal{D}) \label{eq:full_context_score} \\
    S^{-}(y \mid \mathcal{D}) &= \min_{d_i \in \mathcal{D}^+} \sum_{t=1}^{T} \log P(y_t \mid y_{<t}, q, \mathcal{D} \setminus \{d_i\}) \label{eq:loo_score}
\end{align}
Based on these scores, we define the Evidential Contribution $\mathcal{E}(y)$ as the reduction in log-likelihood when this most critical document is removed:
\begin{equation}
\label{eq:evidential_contribution}
    \mathcal{E}(y) = S(y \mid \mathcal{D}) - S^{-}(y \mid \mathcal{D})
\end{equation}
A higher $\mathcal{E}(y)$ indicates that the response $y$ is strongly reliant on a specific piece of evidence from the supporting documents, thus reflecting a higher degree of groundedness.

\subsection{Contrastive Likelihood Reward}
Our sequence-level metric, Evidential Contribution $\mathcal{E}(y)$, is an aggregation of token-level effects. To construct a robust reward signal, we first decompose $\mathcal{E}(y)$ into its token-level components. We define the token-level Evidential Contribution $\epsilon(y_t)$ as:

\begin{equation}
\label{eq:ig_token}
    \epsilon(y_t) = \log P(y_t \mid y_{<t}, q, \mathcal{D}) - \log P(y_t \mid y_{<t}, q, \mathcal{D} \setminus \{d^* \})
\end{equation}
where $d^*$ is the most critical supporting document identified, which minimizes Eq.~\ref{eq:loo_score}. By this definition, the sequence-level contribution is the sum of these token-level scores: $\mathcal{E}(y) = \sum_{t=1}^{T} \epsilon(y_t)$.
However, using $\mathcal{E}(y)$ directly as a reward signal is problematic due to two inherent properties:
\begin{itemize}
    \item \textbf{Length Bias:} As an accumulation of token-level scores, $\mathcal{E}(y)$ is inherently biased towards longer sequences. A model could exploit this by generating verbose or repetitive content from the supporting documents to artificially inflate the reward, leading to degeneration and training instability.
    \item \textbf{Signal Noise and Variance:} The individual token-level contributions $\epsilon(y_t)$ are rarely zero. Small positive values may represent statistical noise rather than genuine grounding. Conversely, negative values can indicate a conflict between the document's content and the model's parametric knowledge. Overly penalizing such complex cases can distract the model from learning from high-quality positive examples.
\end{itemize}
To mitigate these issues, we propose the \textbf{Contrastive Likelihood Reward} ($\mathcal{R}_{CLR}$), which incorporates length normalization and a significance threshold $\tau$:
\begin{equation}
\label{eq:clr_reward}
R_{CLR}(y) = \frac{\mathcal{E}(y) \cdot \mathbb{I}(\mathcal{E}(y) > \tau)}{\sqrt{T}}
\end{equation}
First, normalization by $\sqrt{T}$ mitigates the length bias. This sub-linear penalty discourages excessive verbosity while still allowing genuinely informative long responses to receive a substantial reward (see Section~\ref{sec:Interpretability} for a detailed analysis). Second, the indicator function $\mathbb{I}(\mathcal{E}(y) > \tau)$ with a threshold $\tau > 0$ acts as a significance filter (We name this threshold operation later). It prunes out negligible or negative contributions, ensuring that the model is only incentivized by substantial and unambiguous grounding signals. This prevents the model from over-optimizing on noisy, low-contribution rollouts and enhances training stability.

\subsection{Hybrid Reward Integration}
While the Contrastive Likelihood Reward ($R_{CLR}$) promotes faithfulness by incentivizing outputs attributable to supporting documents, it does not inherently guarantee correctness. This can lead to "faithfully wrong" responses, where the model accurately extracts information from an erroneous document. To rectify this, we must steer the model towards generating answers that are both faithful and factually correct. We achieve this by formulating a hybrid reward that integrates $R_{CLR}$ with an accuracy-based reward, $R_{acc}$.

A key challenge in this integration is that the two rewards operate on different scales. $R_{CLR}$ is an unbounded score with high variance, whereas $R_{acc}$ is typically a binary score. To ensure they are comparable, we first normalize $R_{CLR}$ within each batch of candidate responses. Specifically, we apply group-wise min-max scaling to rescale its values to the range $[0, 1]$:
\begin{equation}
    R'_{CLR}(y_i) = \frac{R_{CLR}(y_i) - \min_{y \in \mathcal{Y}} R_{CLR}(y)}{\max_{y \in \mathcal{Y}} R_{CLR}(y) - \min_{y \in \mathcal{Y}} R_{CLR}(y) + \epsilon}
\label{eq:norm_clr}
\end{equation}
where $\mathcal{Y}$ represents the batch of candidate responses, and $\epsilon$ is a small constant added for numerical stability.

Beyond traditional weighted-sum combinations, we further explore a gating formulation that uses $R'_{\text{CLR}}$ to modulate the influence of $R_{\text{acc}}$:
\begin{equation}
    R_{\text{hybrid}} = R'_{CLR} \cdot R_{\text{acc}}.
\label{eq:hybrid_mul}
\end{equation}
Unlike weighted summation—where even an incorrect but faithful answer may receive partial credit—the gating design assigns zero reward whenever the answer is wrong, thereby placing stronger emphasis on correctness. We adopt this formulation and provide an empirical comparison in Section~\ref{sec:exp_ablation}.

\begin{figure*}[ht]
  \centering

   \includegraphics[width=\textwidth]{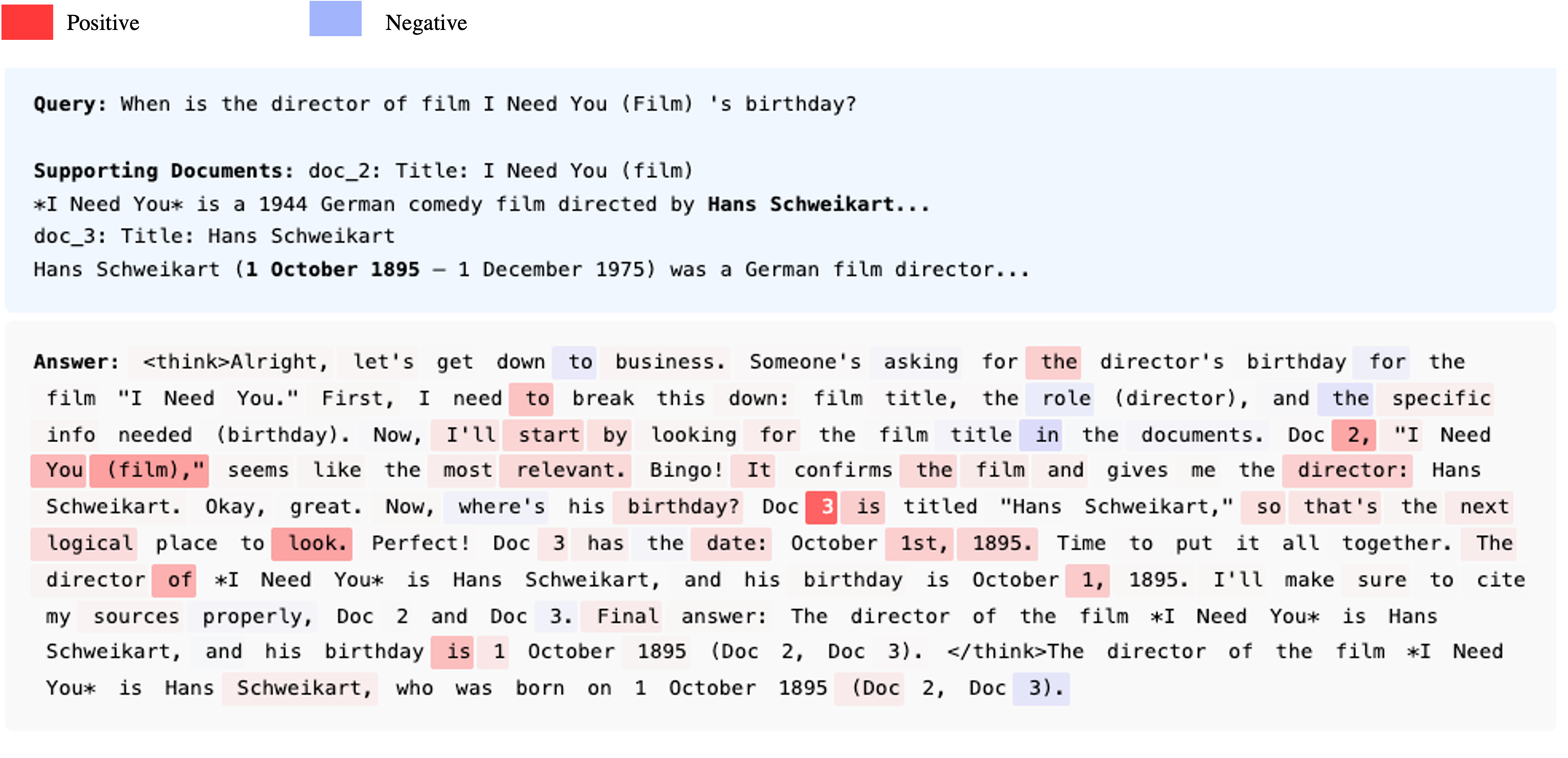}
   \caption{
        An example of token-level Evidential Contribution. The darker the color, the larger the absolute value of $IG_{\text{token}}(y_t)$.
    }
    \label{fig:token-hot}
\end{figure*}
\subsection{Interpretability Analysis: What the Model Learns}
\label{sec:Interpretability}
In the previous sections, we defined our contrastive likelihood-based reward $R_{CLR}$, and proposed two hybrid reward formulations, $R_{\text{hybrid}}$, which integrate $R_{CLR}$ with the accuracy reward $R_{acc}$. In this section, we provide a more intuitive analysis of these rewards to better understand what the model has learned.

We visualize the token-level Evidential Contribution in Figure\ref{fig:token-hot} to examine which tokens are encouraged by our reward model, $R_{CLR}$. From the figure, we observe the following patterns:

\textbf{Grounding in Evidence:} Tokens originating from supporting documents, such as document IDs (e.g., ``2'', ``3'') and their specific contents (e.g., ``1st, 1895''), consistently receive high rewards. This confirms that our method effectively steers the model to locate and utilize relevant source material.

\textbf{Reasoning Linkage:} Connective phrases that forge logical links between disparate pieces of information, like ``so that's the next logical'', are also rewarded. This demonstrates that $R_{CLR}$ encourages the model to synthesize evidence from multiple documents, thereby enhancing both its multi-hop reasoning capabilities.

\textbf{Discouraging Redundancy:} The token-level reward naturally decreases when content is repeated. For example, repeated citations such as ``Doc 2, Doc 3'' receive much lower scores when they reappear later in the answer. This diminishing effect, induced by the $\frac{1}{\sqrt{T}}$ normalization, encourages the model to introduce new information rather than restating the same evidence. As a result, the training becomes more stable, and the final responses are more concise.

Collectively, these observations provide compelling evidence that $R_{CLR}$ yields substantial improvements in document-based reasoning, faithfulness, and efficiency.

\section{Experiment}
\textbf{Datasets.}  To evaluate the document utilization and faithfulness of $R_{CLR}$, we use RAGQALeaderboard \cite{RagQALeaderboard} and PRGB \cite{tan2025prgbbenchmarkrobustplaceholderassisted}.
RAGQALeaderboard includes multi-hop QA tasks—2Wiki \cite{ho2020constructingmultihopqadataset}, HotpotQA \cite{yang2018hotpotqadatasetdiverseexplainable}, and MuSiQue \cite{trivedi2022musiquemultihopquestionssinglehop}—as well as single-hop QA tasks—TriviaQA \cite{joshi2017triviaqalargescaledistantly}, PopQA \cite{mallen2023trustlanguagemodelsinvestigating}, and a biomedical dataset PubMed\cite{jin2019pubmedqadatasetbiomedicalresearch}.PRGB isolates internal model knowledge via placeholders, enabling a more direct assessment of factual faithfulness.
We report accuracy as the primary evaluation metric for all benchmarks.

\textbf{Baselines.} We use Qwen3-8B-Base (Dense) and Qwen3-30B-A3B-Base (MoE) as our base models to demonstrate the robustness of our method across different architectures. Our approach is compared against two main sets of baselines. The first set provides a controlled comparison, starting from a common Supervised Fine-Tuned (SFT) checkpoint: it includes the SFT model itself and models trained using conventional rewards ($R_{\text{acc}}$, $R_{\text{cite}}$). We also evaluate their combination $R_{\text{total}}$, for which we set $\alpha=0.5, \beta=0.5$, and $\eta=0$ according to Equation~\ref{total}. The second set consists of strong, publicly available instruction-tuned models, such as Qwen3-30B-A3B-Instruct-2507 and Qwen3-235B-A22B-Instruct-2507. This allows us to benchmark our model's competitiveness against the state-of-the-art, despite differences in training data.

\textbf{Training and Evaluation .} We train our models using the GRPO algorithm within the ms-swift framework. The training data is constructed from HotpotQA and MuSiQue. For evaluation, we follow the standard prompting and evaluation protocols of RAGQALeaderboard and PRGB. We use a context of 30 documents for RAGQALeaderboard and 10 documents for PRGB. Additional details on training and evaluation are provided in Appendix \ref{sec:train} and \ref{sec:eval}

\subsection{Main Experiments}

\begin{table*}[t]
    \centering
    \caption{\textbf{Overall performance}. The best and second-best results in each group are marked in \textbf{bold} and with an \underline{underline}, respectively. The overall best result is highlighted in \colorbox{mygreen}{green}.}
    \label{tab:my_results_table}
    
    {\small
    \begin{tabular}{llcccccccc}
        \toprule
        \multirow{2}{*}{\textbf{Method}} & \multirow{2}{*}{\textbf{Eval Type}} & \multirow{2}{*}{\textbf{AVG}} & \multicolumn{3}{c}{\textbf{Multi-hop}} & \multicolumn{2}{c}{\textbf{Single-hop}} & \textbf{Faithful} & \textbf{Biomedical} \\
        
        \cmidrule(lr){4-6} \cmidrule(lr){7-8} \cmidrule(lr){9-9} \cmidrule(lr){10-10}
        
        & & & \textbf{2WikI} & \textbf{Hotpot} & \textbf{MuSiQue} & \textbf{TQA} & \textbf{PQA} & \textbf{PRGB} & \textbf{PubMed} \\
        \midrule
        
        \multicolumn{10}{l}{\textit{Qwen3-8B}} \\
        SFT      & \multirow{6}{*}{think}     &     81.6      & 82.0   &     79.6   &     66.8   & 91.4   &     89.6   & 84.7    & 77.2    \\
        SFT+RL($R_{acc}$)    &     &     83.6      & 84.9   &   81.5     &    70.5    & 92.6   &     91.1   & 86.7    & 78.0     \\
        SFT+RL($R_{cite}$)    &   &82.4  &     82.6      & 79.7  &   68.1    &   91.8   & 91.6   &     85.8   & 77.2      \\
        SFT+RL($R_{total}$) &  & \underline{83.9}      &    \underline{85.0} &    81.3    &   \underline{ 70.7 }   &   \underline{92.6}  &  91.4     & 86.6    & \colorbox{mygreen}{\textbf{80.0}}   \\
        SFT+RL($R_{CLR}$)     &    &      83.8     & 84.4  &     \underline{81.6}   &   69.3     & 92.3  &   \underline{92.5}     & \underline{88.1}     & 78.6      \\
        SFT+RL($R_{hybrid}$)  &     &   \textbf{85.0}   & \textbf{86.2}   &    \textbf{82.4}    &    \textbf{71.5}    & \textbf{92.8}   &   \textbf{93.2}     & \textbf{89.4}    & \underline{79.8}   \\

        \midrule 
        \multicolumn{9}{l}{\textbf{Qwen3-8B}} \\
        SFT      & \multirow{6}{*}{ no think}     &     72.0      & 74.6   &    76.1    &   60.2     & 87.7   &   88.6     & 74.1      & 43.0       \\
        SFT+RL($R_{acc}$)    &     &      77.0     & 79.8   & 79.6       &  64.3      & 91.1   &  89.7      & 76.6      & 58.2     \\
        SFT+RL($R_{cite}$)   &     &      77.2     & 78.3   & 78.8       &  62.9      & \underline{91.8}   &  \underline{90.4}      & 74.9     & 63.2    \\
        SFT+RL($R_{total}$) &   &  79.0  &   78.9  &    78.6    &    \underline{64.8}    &   91.6  &   \underline{90.4}    & 76.4    & 74.8   \\
        SFT+RL($R_{CLR}$)      &    &    \underline{79.7}       & \underline{80.4}   &    \underline{79.9}    &    64.4    & 91.6   &    89.6     & \underline{76.6}   & \textbf{75.6}   \\
        SFT+RL($R_{hybrid}$)   &    &    \textbf{80.4}   & \textbf{82.1}   &    \textbf{80.7}    &     \textbf{65.2}   & \textbf{92.1}  &     \textbf{90.7}   & \textbf{77.6}    & \underline{74.8}    \\

        \midrule 
        
        \multicolumn{9}{l}{\textbf{Qwen3-30B-A3B}} \\
        SFT  & \multirow{6}{*}{ think}   &     80.3     & 81.1   &    79.3  &     64.9   & 92.0   & 92.2       & 84.5      & 68.2   \\
        SFT+RL($R_{acc}$)     &    &    83.1     &   85.4 &    82.3    &   70.6     & 93.2   &  93.8     & 85.9      & 70.4       \\
        SFT+RL($R_{cite}$)   & & 82.4    &   83.4  &   80.4     &   69.5     &   93.3  &    94.0    & 85.2    & 70.8 \\
        SFT+RL($R_{total}$)   &  & 84.2      &  86.5   &    \underline{82.5}    &      \underline{73.2}  & 93.7    &    \colorbox{mygreen}{\textbf{95.4}}     & 86.4    & 71.8   \\
        SFT+RL($R_{CLR}$)     &     &      \colorbox{mygreen}{\textbf{85.0}}     & \underline{86.9}  &   \underline{82.5}     &   72.7     & \underline{93.8}   &   94.7     & \colorbox{mygreen}{\textbf{89.5}}    & \textbf{74.6 }   \\
        SFT+RL($R_{hybrid}$)   &  &     \underline{84.9}   &  \colorbox{mygreen}{\textbf{87.1}}   &    \colorbox{mygreen}{\textbf{ 83.2}}    &     \colorbox{mygreen}{\textbf{73.9}}    & \textbf{94.0}   &     \underline{94.9}  & \underline{88.9}      & \underline{72.6}  \\

        \midrule
        \multicolumn{9}{l}{\textbf{Qwen3-30B-A3B}} \\
        SFT   &\multirow{6}{*}{no think}  &      77.6     & 78.3   &    78.7  &     63.4   & 92.3   & 92.2       & 73.8      & 64.2   \\
        SFT+RL($R_{acc}$)     &    &    80.0         &   80.8 &    80.5    &   66.2     & 93.2   &  92.5     & 76.7      & 70.4       \\
        SFT+RL($R_{cite}$)  & & 78.9     &   79.2  &   79.0     &   65.3     &   91.8  &    93.0    & 75.5    & 68.4  \\
        SFT+RL($R_{total}$) & & 81.2     &  82.2   &    81.1    &    \underline{69.1}    &  92.5   &    93.4    & 78.3    & 71.6   \\
        SFT+RL($R_{CLR}$)    &    &   \underline{82.3}   &  \underline{82.7}  &   \underline{81.5}     &   69.0     & \underline{93.7}   &   \underline{93.5}     & \underline{78.7}      & \underline{76.8}   \\
        SFT+RL($R_{hybrid}$)   & &     \textbf{83.1}  & \textbf{83.0}   &    \textbf{81.9}    &    \textbf{70.5}    &  \colorbox{mygreen}{\textbf{94.9}}   &     \textbf{94.4}  & \textbf{79.3}      & \textbf{78.0}   \\

        \midrule
        Qwen3-8B & \multirow{5}{*}{no think}& 61.9 & 71.4 & 71.0 & 44.5& 90.2 &  74.0& 64.7 & 17.6   \\
        Qwen3-14B & & 63.2& 68.8 & 69.4 & 37.6 & 86.8 &  72.7& 65.1 & 42.0 \\
        Qwen3-32B & & 69.3 & 77.9 & 73.2 & 48.6 & 89.4 &  76.4& 68.3 & 51.6   \\
        Qwen3-30B-A3B-Instruct-2507 & & \underline{75.5} & \underline{81.4} & \underline{81.9} & \underline{55.8} & \textbf{94.2} &  \underline{86.7} & \underline{75.9} & \underline{52.4}   \\
        
        Qwen3-235B-A22B-Instruct-2507 &  &\textbf{78.0}  & \textbf{84.9} & \textbf{82.8} & \textbf{65.3} & \underline{93.8} & \textbf{87.6} & \textbf{78.2} & \textbf{53.2} \\
        \bottomrule
    \end{tabular}%
    }
\end{table*}

Table \ref{tab:my_results_table} summarizes the performance of CLR
and all baselines. For multihop tasks that require document-based reasoning, while the accuracy-based reward $R_{acc}$ represents the most straightforward objective, $R_{CLR}$ achieves superior performance. This aligns with our analysis in Section~\ref{sec:Interpretability}, which indicates that $R_{CLR}$ enhances the reward model's ability to identify bridging terms across multiple documents, thereby boosting its reasoning capabilities.

On the PRGB dataset, which is designed to evaluate faithfulness by 
isolating the model's parametric knowledge, CLR again exhibits a significant edge. In every comparison, models trained with $R_{CLR}$ outperform their counterparts by a margin of over 3 points.

In summary, our experimental results robustly demonstrate the superiority of the proposed $R_{CLR}$ reward. 
It consistently outperforms conventional reward functions, such as $R_{acc}$ and $R_{cite}$, across various model architectures and inference settings. 
Furthermore, a hybrid reward, $R_{hybrid}$, which combines $R_{CLR}$ with $R_{acc}$, also surpasses the $R_{total}$ baseline. 
Notably, our model achieves performance competitive with open-source post-trained models, even on multi-hop reasoning tasks where they are known to excel.

\subsection{Experiment Analysis}

\begin{figure}[ht]
  \centering
   \includegraphics[width=0.5\textwidth]{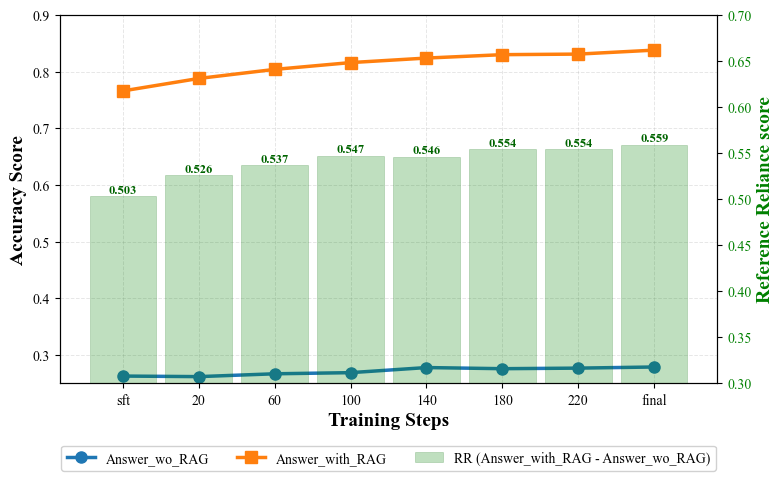}
   \caption{The faithfulness score along with the steps.
    }
    \label{fig:faithful}
\end{figure}

\begin{figure}[ht]
    \centering
    \includegraphics[width=0.5\textwidth]{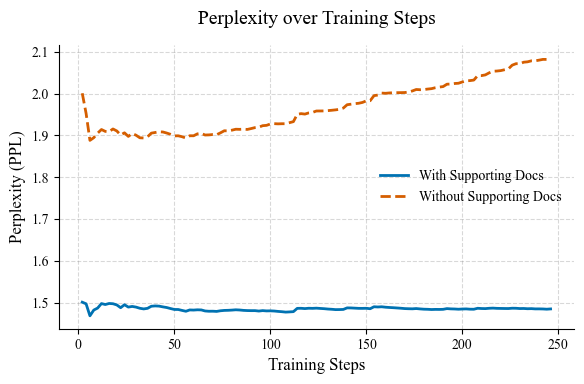}    \caption{Perplexity Length vs. Training Steps}
    \label{fig:perlexity length}
\end{figure}

\textbf{Faithfulness Analysis.} In addition to the PRGB faithfulness benchmark, we define the Reference Reliance score in RagQALeaderboard as $RR_{\theta} = \operatorname{Acc}_{\theta}(Q, D) - \operatorname{Acc}_{\theta}(Q)$,  which can be interpreted as the performance gain when using documents compared to not using them. As shown in Figure~\ref{fig:faithful}, we observe that as the training steps increase, the performance without documents, \(\operatorname{Acc}_{\theta}(Q)\), remains essentially stable, indicating that the model’s internal knowledge does not change significantly. In contrast, the Reference Reliance score \(RR_{\theta}\) improves by 6\%, reflecting a continual strengthening of the model’s ability to leverage external documents effectively. Correspondingly,  increases monotonically.

For the final model, we obtain
\[
\operatorname{Acc}_{\theta}(Q, D) = 0.838,\quad 
\operatorname{Acc}_{\theta}(Q) = 0.279,\quad
RR_{\theta} = 0.559,
\]
compared to Qwen3-30B-A3B-Instruct-2507, for which
\[
\operatorname{Acc}_{\theta}(Q, D) = 0.72,\quad
\operatorname{Acc}_{\theta}(Q) = 0.2912,\quad
RR_{\theta} = 0.429.
\]
From the results compared to baseline above, we can conclude that the CTRL-RAG improves reasoning abilities on documents, along with the improvement of contextual faithfulness.

\textbf{Perplexity Analysis.} The design of our proposed $R_{CLR}$ involves two key scores: the positive score $S(y \mid \mathcal{D})$ and the negative score $S^{-}(y \mid \mathcal{D})$. Analyzing their training dynamics is crucial for understanding the model's learning mechanism. To facilitate this analysis, we track the evolution of their corresponding perplexities, $PPL(y \mid \mathcal{D})$ and $PPL^{-}(y \mid \mathcal{D})$, recalling that $PPL(\cdot) = \exp(-S(\cdot))$.

As illustrated in Figure~\ref{fig:perlexity length}, while $PPL(y \mid \mathcal{D})$ stabilizes after an initial drop, $PPL^{-}(y \mid \mathcal{D})$ steadily rises. This trend signifies a learned shift: the model is moving away from relying on its parametric knowledge and is instead learning to derive answers strictly from the provided documents.

\subsection{Ablation Experiments}
\begin{figure}[ht]
    \centering
    \includegraphics[width=0.5\textwidth]{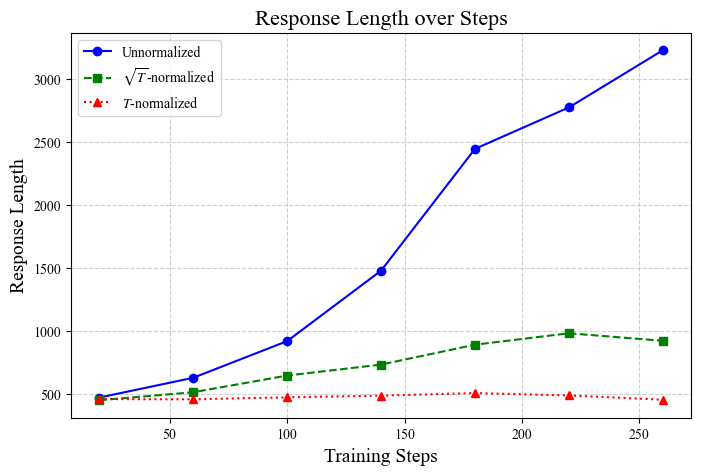}    \caption{Response Length vs. Training Steps}
    \label{fig:response length}
\end{figure}

\textbf{Impact on Length.} 
In our CLR method, to enhance response efficiency from ever-increasing output lengths, we normalize the reward signal by the length of the response.
As illustrated in Figure~\ref{fig:response length}, we experimented with various normalization strategies.
We observed that without any length constraint, the model quickly learns to produce verbose and highly repetitive outputs, a behavior that causes it to consistently hit the maximum generation limit.
Conversely, normalizing the reward directly by the response length~$T$ significantly impedes the learning process, resulting in a static response length. A more effective approach is to normalize by~$\sqrt{T}$.
This method permits an initial increase in response length, which then converges to a stable value, striking a desirable balance between performance and stability.

\textbf{Impact on CLR and its Variants.} 
In the standalone CLR setting, we analyze the leave-one-out (LOO) sequence-level log-likelihood $S^{-}(y \mid \mathcal{D})$. Beyond the minimum selection strategy in Eq.~\ref{eq:loo_score}, we ablate its performance against a standard average pooling:
\begin{equation}
S^{-}(y \mid \mathcal{D}) = \text{avg}_{d_i \in \mathcal{D}^+} \sum_{t=1}^{T} \log P(y_t \mid y_{<t}, q, \mathcal{D} \setminus \{d_i\})
\end{equation}
Results indicate that utilizing the minimum value exploits the "bottleneck effect," guiding the model to better distinguish essential documents for accurate answering, as shown in the upper part of table \ref{tab:ablation}.

\begin{table}[ht]
\caption{Ablation Study Results}
\centering
\label{tab:ablation}
\begin{tabular}{lcc}
\toprule
Ablation Type & RAGQA & PRGB \\
\midrule
SFT & 78.1 & 73.8 \\
\midrule
$\text{LOO}_{\text{min}}$ & 82.1 & 78.3 \\
$\text{LOO}_{\text{avg}}$ & 82.0 & 78.3 \\
$\text{LOO}_{\text{min}}$ with threshold operation & 82.7 & 78.6 \\
$\text{LOO}_{\text{avg}}$ with threshold operation & 82.3 & 78.6 \\
\midrule

$\text{LOO}_{\text{min}}$ with threshold operation + acc & 82.7 & 80.0 \\
$\text{LOO}_{\text{min}}$ with threshold operation * acc & 83.7 & 79.3 \\
$\text{LOO}_{\text{avg}}$ with threshold operation + acc & 83.3 & 79.5 \\
$\text{LOO}_{\text{avg}}$ with threshold operation * acc & 83.1 & 78.2 \\
\bottomrule
\end{tabular}
\end{table}

\textbf{Impact on Integration with Accuracy.} 
\label{sec:exp_ablation}
Building on the min-pooling strategy, we incorporate a correctness signal via an Accuracy metric, which checks for the ground truth in the generated answer (stripping the ``think'' tokens). Comparing multiplicative (Eq.~\ref{eq:hybrid_mul}) and additive fusion, we find that the multiplicative gating approach yields the highest performance gains, as shown in the lower part of table \ref{tab:ablation}.

\subsection{Limitations}
Despite its effectiveness, our framework has several limitations that warrant further investigation. First, the reliance on token-level log-probabilities from the policy model to compute the contrastive reward introduces significant computational overhead during the training phase. Unlike scalar reward models, this approach requires additional forward passes to obtain precise likelihoods, which inevitably reduces training throughput and increases latency. Second, our reward mechanism is designed to prioritize contextual faithfulness, which does not account for knowledge conflicts where the retrieved documents are factually incorrect but the model’s parametric memory is accurate. In such scenarios, the model may be penalized for providing a correct answer that contradicts the erroneous context. Future work will focus on optimizing the efficiency of likelihood estimation and developing adaptive mechanisms to balance parametric trust with contextual adherence when retrieved information is untrustworthy.

\section{Conclusion}
We introduce CTRL-RAG, a novel reinforcement learning framework designed to enhance the faithfulness and contextual reasoning capabilities of the RAG generator. At its core is the Contrastive Likelihood Reward (CLR), a mechanism that synergizes internal model probabilities with external document supervision to explicitly reward grounded generation. This approach effectively steers the model to ground its outputs in provided evidence and sharpen its reasoning abilities. Extensive experiments across both Dense and MoE architectures empirically validate that our method significantly boosts performance in contextual reasoning and faithfulness. These results establish CTRL-RAG as a robust and powerful solution for developing more reliable and context-aware models for complex retrieval-augmented generation tasks.

\newpage


\bibliography{example_paper}
\bibliographystyle{icml2026}

\newpage
\appendix
\onecolumn

\section{Training Settings}
\label{sec:train}
To evaluate the effectiveness of the proposed reward mechanism, we conduct experiments using two distinct architectures of middle size: \textbf{Qwen3-30B-A3B-base} (a Mixture-of-Experts model) and \textbf{Qwen3-8B-base} (a dense model) as our base policy models. The training process follows a two-stage pipeline. Initially, we perform Supervised Fine-Tuning (SFT) to obtain a "cold-start" model. Subsequently, we execute Reinforcement Learning (RL) using the GRPO algorithm. We test two RL configurations: \textbf{CLR-only} (Contrastive Log-likelihood Reward) and \textbf{CLR combined with Accuracy} (a hybrid of CLR and accuracy-based rewards). Our evaluation covers both reasoning and non-reasoning inference modes, with results demonstrating significant performance improvements across all settings.
\textbf{Dataset Composition and Selection.} The training data is curated from a combination of HotpotQA and MusiqueQA. For the SFT phase, we select samples based on the $pass@8$ accuracy metric. The SFT dataset comprises 74,109 samples in total, with 80\% being samples that have $pass@8 = 1.0$ and 20\% being samples with $pass@8 \in [0.5, 0.875]$. After excluding the SFT set, the remaining data serves as the candidate pool for RL. From this pool of approximately 10,000 samples, we select 90\% with $pass@8 \in [0.1, 0.875]$ and 10\% with $pass@8 = 1.0$ to form the final RL training set. To ensure the quality of contrastive learning, we specifically select instances where the standard deviation of the calculated log-probabilities exceeds a threshold of 10 for CLR training.

\begin{figure*}[ht]
  \centering
   \includegraphics[width=\textwidth]{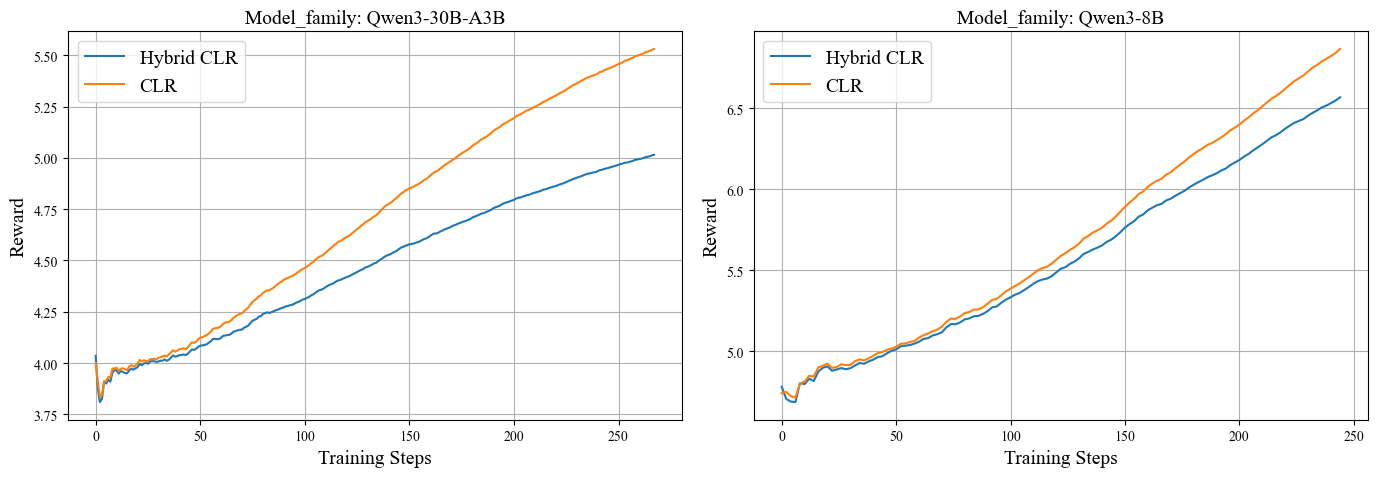}
   \caption{
        Reward Curve.
    }
    \label{fig:reward}
\end{figure*}

\textbf{Implementation Details.} We implement our training framework using MS-Swift, with a customized rewarding service modified to support entropy regularization. The RL stage utilizes a group size of 8, a learning rate of $1 \times 10^{-6}$, and an effective batch size of 1024. The training is distributed across a heterogeneous cluster of $32 \times \text{H800}$ and $8 \times \text{H800}$ GPUs, utilizing Tensor Parallelism (TP=4) and Pipeline Parallelism (PP=4) to support the rollout service and optimization, the reward curve in training is shown in Figure ~\ref{fig:reward}.

\textbf{Omission of KL Divergence.}  Notably, we exclude the KL divergence penalty during the RL process. In the standard GRPO objective, the KL term is defined as $\beta \log \frac{\pi_\theta}{\pi_{ref}}$. However, a fundamental optimization conflict arises in our setup: (1) the CLR reward explicitly encourages the model to increase the probability $\pi_\theta(y |Q, D)$, whereas (2) the KL term penalizes any deviation of $\pi_\theta$ from $\pi_{ref}$. This contradiction makes it difficult for the model to optimize the contrastive reward effectively. In preliminary experiments, we observed that including the KL term resulted in a sharp, continuous decline in reward values; thus, its removal is essential for stable convergence. In our practical experiments, we conducted tests by setting the KL divergence coefficient $\beta$ to $0.05$, $0.04$, $0.03$, and $0.01$. We observed that the training collapsed within $20$ steps in all instances--characterized by reward reduction, garbled rollout outputs, or repetitive generation—which empirically validates the effectiveness of our theoretical framework.

\textbf{Rationale for the Significance Threshold $\tau$.}
The threshold $\tau$ in our Contrastive Likelihood Reward (Equation~\ref{eq:clr_reward}) serves as a crucial filter to distinguish meaningful contextual grounding from statistical noise. 
We determined its value empirically by setting $\tau=1$. 
This decision was informed by a preliminary analysis of our training data, specifically the samples curated via the $pass@8$ selection criterion. 
Our analysis revealed a strong correlation: for generated responses where the Evidential Contribution score $R_{CLR}$ was less than 1, a vast majority (over 95\%) were found to be entirely ungrounded, containing no discernible information from the supporting documents.
This finding suggests that $R_{CLR}$ acts as a reliable heuristic for identifying non-faithful outputs. 
Consequently, we treat these low-score instances uniformly by assigning them a zero reward. This strategy simplifies the reward landscape by filtering out irrelevant rollouts and ensures that the reinforcement learning process concentrates on optimizing for genuinely context-aware and faithful generations.

\section{Evaluation Settings}
\label{sec:eval}
We use benchmark in RagQALeaderboard\cite{RagQALeaderboard}, which provided a fair environment of RAG-QA scenarios to evaluate models, and PRGB\cite{tan2025prgbbenchmarkrobustplaceholderassisted} to evaluate contextual faithfulness of LLMs. For inference, we leverage the vLLM framework deployed on a cluster of eight NVIDIA H800 GPUs. Regarding specific task configurations, we retrieve 30 documents for RagQALeaderboard. For the PRGB benchmark, we conduct three iterations with a noise distribution of 4, 4, and 1 for noise document levels 1, 2, and 3, respectively. 
For Evaluation prompts, we followed the default prompts of their benchmarks.

\section{Detailed Ablation Experiment Results}

\begin{table*}[ht]
\caption{Detailed Ablation Study Results}
\label{tab:results}
\centering
\begin{tabular}{lcccccccc}
\toprule
model & final\_score & musiqueQA & hotpotQA & 2Wiki & PQA & TQA & PubMed & PRGB \\
\midrule
SFT    &      77.6     & 63.4   &    78.7  &   78.3     & 92.2   & 92.3       & 64.2      &    73.8   \\
\midrule
$\text{LOO}_{\text{min}}$ & 81.8 & 68.2 & 81.5 & 81.9 & 93.2 & 93.6 & 75.8 & 78.3 \\
$\text{LOO}_{\text{avg}}$ & 81.7 & 68.2 & 81.4 & 81.9 & 93.2 & 93.5 & 75.6 & 78.4 \\
$\text{LOO}_{\text{min}}$ with threshold operation & 82.3 & 69.0 & 81.5 & 82.7 & 93.5 & 93.7 & 76.8 & 78.6 \\
$\text{LOO}_{\text{avg}}$ with threshold operation & 82.2 & 69.7 & 81.7 & 82.7 & 93.2 & 94.0 & 75.2 & 78.6 \\
\midrule
$\text{LOO}_{\text{min}}$ with threshold operation + acc & 82.9 & 70.9 & 82.6 & 83.9 & 94.2 & 93.7 & 75.2 & 80.0 \\
$\text{LOO}_{\text{min}}$ with threshold operation * acc & 83.1 & 70.5 & 81.9 & 83.0 & 94.4 & 94.9 & 78.0 & 79.3 \\
$\text{LOO}_{\text{avg}}$ with threshold operation + acc & 83.0 & 70.3 & 82.2 & 83.6 & 93.9 & 94.3 & 77.0 & 79.5 \\
$\text{LOO}_{\text{avg}}$ with threshold operation * acc & 82.4 & 69.3 & 81.9 & 82.6 & 93.7 & 93.3 & 77.8 & 78.2 \\
\bottomrule
\end{tabular}
\end{table*}

\end{document}